\documentclass[11pt,a4paper]{article}
\usepackage{times,latexsym,amssymb}
\usepackage{url}
\usepackage{booktabs}
\usepackage[table,xcdraw]{xcolor}
\usepackage{multirow,rotating,tikz}
\usepackage{subcaption}
\usepackage{graphicx}

\def\equationautorefname~#1\null{(#1)\null}
\def\itemautorefname~#1\null{(#1)\null}
\def\sectionautorefname~#1\null{\S#1\null}
\def\subsectionautorefname~#1\null{\S#1\null}

\usepackage[acceptedWithA]{tacl2021v1}
\usepackage[noabbrev,capitalize,nameinlink]{cleveref}
\iftaclpubformat

\else

\fi

\title{Can Authorship Representation Learning Capture Stylistic Features?}

\author{
 Andrew Wang$^{1 *}$,
 Cristina Aggazzotti$^{1 *}$,
 Rebecca Kotula$^2$,
 \\
  {\bf Rafael Rivera Soto$^3$,
  Marcus Bishop$^2$,
  Nicholas Andrews$^{1 \dagger}$}
  \\
  $^1$ Johns Hopkins University,
  $^2$ U.S. Department of Defense,\\
  $^3$ Lawrence Livermore National Laboratory\\
}

\begin{document}
\maketitle
\begin{abstract}
Automatically disentangling an author's style from the content of their writing is a longstanding and possibly insurmountable problem in computational linguistics.
At the same time, the availability of large text corpora furnished with author
labels has recently enabled learning authorship representations in a purely data-driven manner for authorship attribution,
a task that ostensibly depends to a greater extent on encoding writing style than encoding content.
However, success on this surrogate task does not ensure that such representations capture writing style since authorship could also be correlated with other latent variables, such as topic. 
In an effort to better understand the nature of the information these representations convey, and specifically to validate the hypothesis that they chiefly encode writing style,
we systematically probe these representations through a series of targeted experiments.
The results of these experiments suggest that representations learned for the surrogate authorship prediction task are indeed sensitive to writing style.
As a consequence, authorship representations may be expected to be robust to certain kinds of data shift, such as topic drift over time.
Additionally, our findings may open the door to downstream applications that require stylistic representations, such as style transfer.
\end{abstract}
\def\thefootnote{*}\footnotetext{These authors contributed equally to this work.}
\def\thefootnote{\arabic{footnote}}
\def\thefootnote{$\dagger$}\footnotetext{Corresponding author: \href{mailto:noa@jhu.edu}{noa@cs.jhu.edu}}
\def\thefootnote{\arabic{footnote}}

\section{Introduction}\label{sec:intro}Knowing something about an author's writing style is helpful in many applications, such as 
predicting who the author is, determining which passages of a document the author composed, rephrasing text in the style of another author, and generating new text in the style of a particular author.
The trouble is that fully characterizing something as complex as writing style has proven too unwieldy to admit fine-grained human annotations, which leaves the possibility of directly learning explicit and interpretable representations of writing style practically beyond reach.
Instead, research in this area has largely focused on specific stylistic attributes, such as formality, toxicity, politeness, gender, simplicity, and humor, which are more straightforward to annotate~\cite{rao2018dear,pavlopoulos2020toxicity,madaan2020politeness,li2018delete,jin2022deep}. Unfortunately, the reliance on human labels and the narrow focus of such stylistic distinctions severely limit the utility of such representations in tasks related to authorship, such as those listed above.

In this paper, we focus instead on the {\em authorship prediction task}, which enjoys the benefit of not requiring manually-elicited labels, since metadata in many corpora include either explicit author labels or usernames that may serve as proxies for latent authorship.
As a result, the vast scale of data available for training authorship prediction models opens the door to learning {\em generalizable} authorship representations using deep learning. We specifically consider similarity learning approaches that aim to produce vector representations of documents, where the distance between two vectors is inversely related to the likelihood that the corresponding documents were composed by the same author~\cite{boenninghoff2019explainable,andrews2019learning}.  

However, achieving high accuracy in the authorship prediction task does not necessarily imply that stylistic features have been successfully learned.
For example, in a given corpus, correctly predicting that two writing samples were composed by the same author may be possible on the basis of non-stylistic signal, such as the topic of conversation.
Therefore, this work is concerned with obtaining a better understanding of the nature of representations learned for the authorship prediction task.

Unfortunately, because deep learning models behave like black boxes, we cannot directly interrogate a model's parameters to determine what information such representations contain.
For example, one might hope to employ attention-based approaches that provide post hoc explanations through token saliency maps~\citep{sundararajan2017axiomatic}. However, such methods provide no guarantee of the fidelity of their explanations to the underlying model. Furthermore, the subjective interpretation required to deduce the reasons that such methods highlight certain spans of text makes it nearly impossible to systematically draw conclusions about the model.

Instead, we propose targeted interventions to probe representations learned for the surrogate authorship prediction task.
First, we explore masking content words at training time in~\autoref{sec:mask}, an operation intended to gauge the degree to which a representation relies on content.
Then we explore automatic paraphrasing in~\autoref{sec:andrew}, an operation intended to preserve meaning while modifying how statements are expressed.
Finally, in~\autoref{sec:gen} we explore the capacity of these representations to generalize to unseen tasks, specifically topic classification and coarse style prediction.

Taken together, and despite approaching the research question from various points of view, our experiments suggest that representations derived from the authorship prediction task are indeed substantially stylistic in nature. In other words, success at authorship prediction may in large part be explained by having successfully learned discriminative features of writing style. The broader implications of our findings are discussed in~\autoref{sec:discussion}.
\section{Related Work}\label{sec:related}Perhaps the work most closely related to our study is that of~\citet{wegmann-nguyen-2021-capture} and~\citet{wegmann2022same} who propose measuring the stylistic content of authorship representations through four specific assessments, namely formality, simplicity, contraction usage, and number substitution preference.
Our work differs in two main  respects.
First, we regard style as an abstract constituent of black-box authorship representations rather than the aggregate of a number of specific stylistic assessments.
Second, the works above deal with stylistic properties of {\em individual sentences}, whereas we use representations that encode longer spans of text.
Indeed, we maintain that the writing style of an author manifests itself only after observing a sufficient amount of text composed by that author.
For example, it would be difficult to infer an author's number substitution preferences after observing a single sentence, which is unlikely to contain multiple numbers. The same is true of other stylometric features, such as capitalization and punctuation choices, abbreviation usage, and characteristic misspellings.

In another related work, \citet{sari-etal-2018-topic}
find that although content-based features may be suitable for datasets with high topical variance,
datasets with lower topical variance benefit most from style-based features.
Like the works mentioned above, \citeauthor{sari-etal-2018-topic} explicitly identify a number of style-based features, so writing style is more of a premise than the object of study.
In addition, experiments in this previous work are limited to datasets featuring a small number of authors, with the largest dataset considered containing contributions of only 62 authors.

A number of end-to-end methods have been proposed to learn representations of authorship~\cite{andrews2019learning,boenninghoff2019explainable,saedidras2021,hay2020,huertas-tato}. A common thread among these approaches is their use of {\em contrastive learning}, although they vary in the particular objectives used. They also differ in the the domains used in their experiments, the numbers of authors considered for training and evaluation, and their open- or closed-world perspectives.
As discussed in~\autoref{sec:author_rep}, we use the construction introduced in~\citet{Soto2021LearningUA} as a representative neural method because it has shown evidence of capturing stylistic features
through both its success in the challenging open-world setting in multiple domains and its performance in zero-shot domain transfer.
\section{Authorship Representations}\label{sec:author_rep}In this article, an {\em authorship representation} is a function mapping documents to a fixed Euclidean space.
The fact that such representations are useful for a number of authorship-related tasks is generally attributed to their supposed ability to encode author-specific style, an assertion we aim to validate in this paper.
In this section, we describe how these representations arise and how they are intended to be used.

Our analysis centers around representations~$f$ implemented as deep neural networks and trained using a {\em supervised contrastive objective}~\cite{khosla2020supervised}. At training time this entails sampling pairs of documents $x,x'$ composed by the same author (resp. by {\em different} authors) and minimizing (resp. {\em maximizing}) the distance between $f\left(x\right)$ and $f\left(x'\right)$. Therefore, we may assume at inference time that $f\left(x\right), f\left(x'\right)$ are closer together if $x,x'$ were composed by the same author than they would be if $x,x'$ were composed by different authors.
No meaning is ascribed to any attribute of $f\left(x\right)$, such as its coordinates, its length, or its direction. Rather, $f\left(x\right)$ is meaningful only in relation to other vectors.

In all the experiments of this paper $f$ is an instance of the {\em Universal Authorship Representation} (UAR) introduced in~\citet{Soto2021LearningUA}.\footnote{\noindent An open-source implementation is available at \url{https://github.com/LLNL/LUAR}.}
Notwithstanding the merits of a number of other recent approaches discussed in~\autoref{sec:related}, we argue that because of its typical neural structure and typical contrastive training objective, UAR serves as a representative model. 
The same paper also illustrates that UAR may be used for zero-shot transfer between disparate domains,
suggesting a capacity to learn generalizable features, perhaps of a stylistic nature, thereby making it a good candidate for our experiments.

Note that UAR defines a {\em recipe} consisting of an architecture and a training process that must be carried out in order to arrive at a representation, with the understanding that care must be taken in assembling appropriate training datasets.
Specifically, we consider a diverse set of authors at training time in an effort to promote representations that capture \emph{invariant} authorship features, chiefly writing style, rather than time-varying features, such as topic.
Invariance is a desirable feature of authorship representations because it improves the likelihood of  {\em generalization} to novel authors or even to novel domains. However, there is no guarantee that invariant features are exclusively stylistic in any given corpus, or that any training process we might propose will result in representations capturing exclusively invariant features. Therefore, this work is concerned with estimating the \emph{degree} to which authorship representations are capable of capturing stylistic features, with the understanding that completely disentangling style from topic may be beyond reach.
\section{Experimental Setup}\label{sec:setup}Mirroring~\citet{Soto2021LearningUA}, we conduct experiments 
involving three datasets, each consisting of documents drawn from a different domain.
For the reader's convenience, we present further details and some summary statistics of these datasets in~\autoref{sec:dataAppendix}.

To evaluate an authorship representation, we use the common experimental protocol described below.
The objective is to use the representation to retrieve documents by a given author from among a set of candidate documents, which are known as the {\em targets}, on the basis of the distances between their representations and the representation of a document by the desired author, which is known as the {\em query}.
To this end, each evaluation corpus has been organized into queries and targets, which are used to calculate the \emph{mean reciprocal rank (MRR)}.
Following is a friendly description of this metric, with a more elaborate formulation presented in~\autoref{sec:mrrAppendix}.

An authorship representation may be used to sort the targets according to the distances between their representations and that of any fixed query.
In fact, this ranking is often seen as the primary outcome of an authorship representation.
Because one would need to manually inspect the targets in the order specified by the ranking, it would be desirable for any target composed by the same author as the query to appear towards the beginning of this list.
The MRR is the expectation of the reciprocal of the position in the ranked list of the first target composed by the same author as a randomly chosen query.
This metric ranges from 0 to 1, with higher values indicating a greater likelihood of finding documents composed by an author of interest within a large collection in the first few search results.

Following~\citeauthor{Soto2021LearningUA} the queries and targets in all our experiments are {\em episodes}, each consisting of 16~comments or product reviews contiguously published by the same author in the Reddit or Amazon domains, respectively, or 16 contiguous paragraphs of the same story in the fanfic domain.

In order to conduct a wide variety of experiments in a time-efficient manner, we train all representations on one GPU for 20 epochs,
although we acknowledge that better results may be obtained by training with more data, on multiple GPUs, or for longer than 20 epochs.
\section{Masking Content Words}\label{sec:mask}Our first series of experiments aims to illustrate through a simple training modification that authorship representations are capable of capturing style.
Specifically, the strategy of {\em masking} training data in a way that preserves syntactic structure, something which is known to relate to style, while removing thematic or topical information, has been effective, particularly in cross-domain authorship experiments~\citep{stamatatos2018}.
To this end, we propose training authorship representations with restricted access to topic signal by masking varying proportions of content-related words in the training data.
Evaluating each of these representations and comparing its ranking performance with that of a representation trained on the same {\em unmasked} data reveals the capacity of the representation to capture style.

Words may be roughly divided into two categories: {\em content words} and {\em function words}.
Content words primarily carry topic signal. They tend to include nouns, main verbs, adjectives, and adverbs.
Function words serve syntactic roles and convey style through their patterns of usage.
They tend to include auxiliary verbs, prepositions, determiners, pronouns, and conjunctions~\citep{mostellerwallace}.
These observations suggest masking words according to their parts of speech (POS), a process we call {\em Perturbing Linguistic Expressions} or {\em PertLE}.

\subsection{The PertLE schema} \label{sec:pertle}
In our PertLE masking schema, we replace all words belonging to certain POS categories with a distinguished masking token.
This approach stems from the observation that content words may often be distinguished from function words on the basis of POS. However, this is simply a heuristic and there are many exceptions.
For instance, although many adverbs may be categorized as content words, such as~\emph{happily}, others play a functional role, such as~\emph{instead}. 
Because masking on the basis of POS is an imperfect strategy to eliminate content, we introduce the following  {\em levels} of the PertLE schema. In our {\em PertLE Grande} schema we mask all nouns, main verbs, adjectives, and adverbs. This is a greedy approach intended to mask words that could possibly convey content, at the expense of occasionally masking some function words.
In contrast, in our {\em PertLE Lite} schema we mask only nouns, which are most likely to carry content information.\footnote{
We also tried masking every word belonging to certain POS categories with a distinguished {\em pseudoword} specific to its POS. These pseudowords were selected to be morphologically similar to other words in their POS categories but not appear in our corpora. However, we adopt the simpler masking approach described above because it surprisingly produced very similar ranking results.}
In a follow-up reported in~\autoref{sec:tertle} we repeat the main experiment below using a masking schema based on TF-IDF scores rather than POS.

\paragraph{Procedure}
To identify POS categories, we use the Stanford NLP Group's Stanza tokenizer and POS tagger \citep{qi2020stanza} due to their efficiency, flexibility, versatility, and capacity for handling other languages.
We use the Universal POS (UPOS) tagset because it distinguishes between main verbs (VERB) and auxiliary verbs (AUX), labels \emph{not} and \emph{-n't} as particles rather than adverbs, and tags many foreign language words with their correct POS category rather than labeling them as foreign words.
For both masking levels, we replace each word to be masked with SBERT's masking token, \verb!<mask>!,
preserving any contracted particles (e.g.~\verb!gonna! $\rightsquigarrow$ \verb!<mask>na!).
As an example, \autoref{fig:example} illustrates both levels of the PertLE schema applied to the same statements.

\begin{figure}[hb!]
\label{fig:pertleExample}\scriptsize
{\bf Unmasked:}
\begin{itemize}\setlength\itemsep{0em}
\item Hold me closer, tiny dancer. Count the headlights on the highway. Lay me down in sheets of linen. You had a busy day today.
\item Just a small-town girl, livin' in a lonely world. She took the midnight train going anywhere.
\item All I wanna do is have a little fun before I die, says the man next to me out of nowhere.
\end{itemize}
{\bf PertLE Grande:}
\begin{itemize}\setlength\itemsep{0em}
\item \verb!<mask>! me \verb!<mask>!, \verb!<mask>! \verb!<mask>!. \verb!<mask>! the \verb!<mask>! on the \verb!<mask>!. \verb!<mask>! me \verb!<mask>! in \verb!<mask>! of \verb!<mask>!. You \verb!<mask>! a \verb!<mask>! \verb!<mask>! \verb!<mask>!.
\item \verb!<mask>! a \verb!<mask>! \verb!<mask>!, \verb!<mask>! in a \verb!<mask>! \verb!<mask>!. She \verb!<mask>! the \verb!<mask>! \verb!<mask>! \verb!<mask>! \verb!<mask>!.
\item All I \verb!<mask>!na \verb!<mask>! \verb!<mask>! \verb!<mask>! a \verb!<mask>! \verb!<mask>! before I \verb!<mask>!, \verb!<mask>! the \verb!<mask>! \verb!<mask>! to me out of \verb!<mask>!.
\end{itemize}
{\bf PertLE Lite:}
\begin{itemize}\setlength\itemsep{0em}
\item Hold me closer, tiny \verb!<mask>!. Count the \verb!<mask>! on the \verb!<mask>!. Lay me down in \verb!<mask>! of \verb!<mask>!. You had a busy \verb!<mask>! \verb!<mask>!.
\item Just a small-town \verb!<mask>!, livin' in a lonely \verb!<mask>!. She took the \verb!<mask>! \verb!<mask>! going anywhere.
\item All I wanna do is have a little \verb!<mask>! before I die, says the \verb!<mask>! next to me out of nowhere.
\end{itemize}
\caption{\small Various levels of the PertLE masking schema applied to the same statements.}
\label{fig:example}
\end{figure}

Using the procedure described in~\citet{Soto2021LearningUA}, for each domain we train multiple authorship representations on that domain's training corpus: one with the training corpus masked according to PertLE Grande, one masked according to PertLE Lite, and one unmasked to serve as a baseline.
We evaluate each representation on each {\em unmasked} evaluation corpus to afford a fair comparison of the effects of the masking level for each combination of training and evaluation domain.

Note that for representations trained on {\em masked} data, this evaluation introduces a {\em mismatch} between training and evaluation datasets, although the baseline representations remain unaffected.
In cases where masking results in a large {\em degradation} in performance, this setup makes it impossible to distinguish between our interventions and the train-test mismatch as the cause of the degradation.
On the other hand, this distinction is immaterial in the case that masking does {\em not} degrade performance, and in fact, this case is the desired outcome of the experiment, as it would suggest that the corresponding representation does not benefit significantly from the information withheld by masking content words.

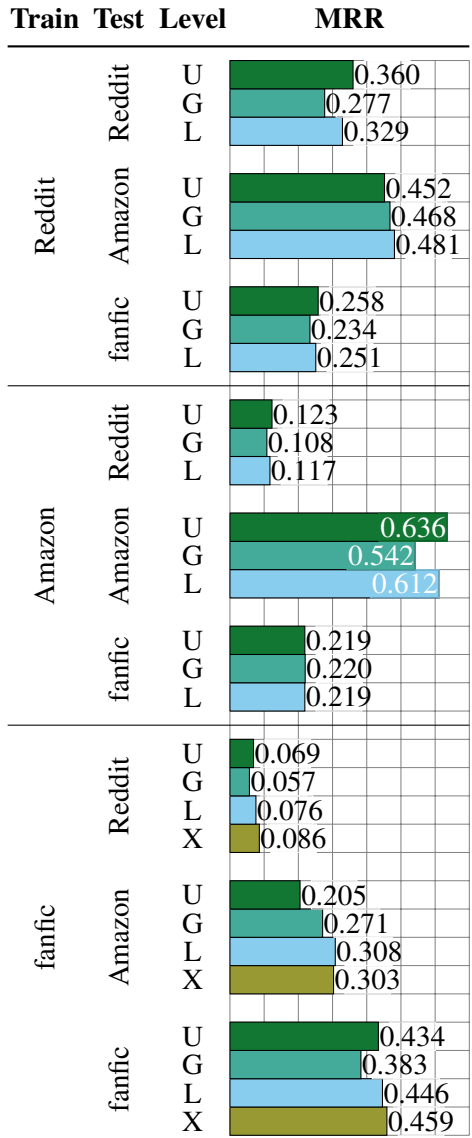
\begin{figure}[hb!]
\begin{center}
\definecolor{c0}{RGB}{51,34,136}
\definecolor{c1}{RGB}{136,204,238}
\definecolor{c2}{RGB}{68,170,153}
\definecolor{c3}{RGB}{17,119,51}
\definecolor{c4}{RGB}{153,153,51}
\begin{tikzpicture}[scale=1.5]

\draw[xstep=.3,ystep=0.25,gray,very thin] (0,0) grid (2.1,9.5);
\draw (-1.95,3.625) -- (2.1,3.625);
\draw (-1.95,6.625) -- (2.1,6.625);

\foreach \l/\y/\z in { fanfic/0.00/3.5, Amazon/3.75/6.50, Reddit/6.75/9.50 }
  \draw[white] (-1.95,\y) rectangle (-1.3,\z) node[pos=0.5,black] {\begin{sideways}\l\end{sideways}};

\foreach \l/\y/\z in {
fanfic/00.00/1.00, Amazon/1.25/2.25, Reddit/02.50/3.50,
fanfic/03.75/4.50, Amazon/4.75/5.50, Reddit/05.75/6.50,
fanfic/06.75/7.50, Amazon/7.75/8.50, Reddit/08.75/9.50 }
  \draw[white] (-1.3,\y) rectangle (-0.65,\z) node[pos=0.5,black] {\begin{sideways}\l\end{sideways}};

\foreach \x/\c/\y/\l in {
0.459/c4/00.00/X, 0.446/c1/00.25/L, 0.383/c2/00.50/G, 0.434/c3/00.75/U,
0.303/c4/01.25/X, 0.308/c1/01.50/L, 0.271/c2/01.75/G, 0.205/c3/02.00/U,
0.086/c4/02.50/X, 0.076/c1/02.75/L, 0.057/c2/03.00/G, 0.069/c3/03.25/U,
0.219/c1/03.75/L, 0.220/c2/04.00/G, 0.219/c3/04.25/U,
0.117/c1/05.75/L, 0.108/c2/06.00/G, 0.123/c3/06.25/U,
0.251/c1/06.75/L, 0.234/c2/07.00/G, 0.258/c3/07.25/U,
0.481/c1/07.75/L, 0.468/c2/08.00/G, 0.452/c3/08.25/U,
0.329/c1/08.75/L, 0.277/c2/09.00/G, 0.360/c3/09.25/U
}{
  \draw[white] (-0.65,\y) rectangle (0,\y+0.25) node[pos=0.5,black]{\l};
  \draw[fill=\c] (0,\y) rectangle (\x*3,\y+0.25);
  \draw[white] (\x*3+.1,\y) rectangle (\x*3+0.5,\y+0.25) node[pos=0.5,black]{\x};
}

\foreach \x/\c/\y/\l in {
0.612/c1/04.75/L, 0.542/c2/05.00/G, 0.636/c3/05.25/U
}{
  \draw[white] (-0.65,\y) rectangle (0,\y+0.25) node[pos=0.5,black]{\l};
  \draw[fill=\c] (0,\y) rectangle (\x*3,\y+0.25);
  \draw[\c] (\x*3-0.6,\y) rectangle (\x*3,\y+0.25) node[pos=0.5,white]{\x};
}
\draw[thick] (-1.95,9.625) -- (2.1,9.625);
\draw[white] (-1.95,9.75) rectangle (-1.3,10.00) node[pos=0.5,black] {\bf Train};
\draw[white] (-1.3,9.75) rectangle (-0.65,10.00) node[pos=0.5,black] {\bf Test};
\draw[white] (-.65,9.75) rectangle (0,10.00) node[pos=0.5,black] {\bf Level};
\draw[white] (0,9.75) rectangle (2.1,10.00) node[pos=0.5,black] {\bf MRR};

\end{tikzpicture}
\caption{\small MRR results for models trained on unmasked data (U), or data masked according to the PertLE Grande (G), the PertLE Lite (L), or additionally for fanfic, the PertLE Xtra-Lite (X) schema.} 
\label{table:pertle}
\end{center}
\end{figure}

The results of the experiment are shown in \autoref{table:pertle}.
For each corpus and each masking level, we independently trained {\em three} representations in an effort to reduce variance.
Each number reported is the sample mean of the MRR according to each of the three independent representations, where 0.014 is the maximum sample standard deviation over all experiments reported in the figure.

\paragraph{Discussion}
A one-way ANOVA was performed for each combination of training and evaluation domain, showing that the masking schema had a statistically significant impact on the mean values of the MRR reported in~\autoref{table:pertle} with $p<0.01$ in all cases except the case with training domain Amazon and evaluation domain fanfic. In that case, we conclude that masking words at training time had no significant effect on ranking performance, as desired. In the other cases the change in performance {\em was} significant but relatively minor.

In cases where performance {\em improved}, we believe the most likely explanation is that, deprived of content words at training time, the model was forced to discover other authorship features, which turned out to be more useful than content words in the corresponding evaluation domains. In cases where performance {\em dropped}, it appears that the model was unable to compensate for the loss of content words. However, we emphasize that in these cases the drop in performance is surprisingly small in light of the fact that PertLE Grande masks nearly {\em half} of all training tokens. See~\autoref{table:percentages} in~\autoref{sec:tertle} for the exact proportions of tokens masked in each domain or \autoref{fig:example} for a qualitative example. Another possibility is that, as discussed above, PertLE Grande obscures writing style to some extent, which could also account for the small drop in ranking performance if the representations were primarily style-focused.

For all three training and evaluation domains, the MRR of the Lite model is quite close to that of the unmasked model. 
This suggests that masking words most likely to convey content changes ranking performance very little, and even improves it in some settings.
We also observe that although the MRR of each Grande model is generally less than that of the corresponding
Lite model, it is not dramatically so.
This suggests that increasing the proportion of tokens masked appears to eventually impair ranking performance, but not to the degree one might expect given the considerable proportion of words masked.

We know of no way to {\em completely} redact the content of a document while retaining its writing style.
We doubt that this is even possible, least of all in an automated fashion. It follows that the representations trained on data masked according to the PertLE schema (as well as the TertLE schema discussed~\autoref{sec:tertle}) probably {\em do} encode a small amount of content. Being trained to distinguish authors on the basis of such masked text, these models are therefore likely to learn to use that information to their advantage when appropriate, which would mean that the representations considered in this paper {\em do} convey a small amount of topical signal, an observation which is corroborated by the experiments in~\autoref{sec:andrew} and \autoref{sec:gen}.

Nevertheless, the experiment shows that PertLE obscures {\em much} of the content of a training corpus, which in turn affects ranking performance only marginally. We argue that those representations are therefore likely to have learned to avail of features other than content, thereby illustrating their {\em capacity} to avail of writing style.

\subsection{PertLE Xtra-Lite}\label{sec:XtraLite}

As observed in~\citet{Soto2021LearningUA} the representation trained on the fanfic corpus generalizes poorly to the other two domains, something which is probably due to the comparatively small size and lack of topical diversity of that dataset.
This suggests that representations trained on fanfic stand to improve the most by a targeted inductive bias.
Indeed, the Lite model trained on the fanfic dataset improves performance in all three evaluation domains.
This may be explained by the observation that the fanfic domain may contain more jargon and specialized language appearing in the form of proper nouns representing names, places, and things. 
This is borne out by the observation that in the Reddit, Amazon, and fanfic domains, around 22\%, 20\%, and 35\% respectively of all nouns are proper. 

To further explore this observation we introduce the {\em Xtra-Lite} level of the PertLE schema, in which we mask only proper nouns, the POS category most likely to convey content information.
Repeating the same procedure as before, we train a PertLE Xtra-Lite (X) representation on the fanfic domain and evaluate it on each unmasked evaluation dataset.
The results in~\autoref{table:pertle} show that
the Xtra-Lite model not only outperforms the unmasked and Grande models in all three domains, but also outperforms the Lite model in the Reddit and fanfic domains and performs nearly as well in the Amazon domain, confirming that representations trained on fanfic benefit from a targeted inductive bias. 

\section{Removing Style by Paraphrasing}\label{sec:andrew}In contrast with the experiments in~\autoref{sec:mask} that aim to assess the ability of authorship representations to capture style by removing  {\em content}, our next group of experiments aims to make the same assessment by instead removing {\em style}.
For this purpose we consider automatic paraphrasing, which ideally introduces stylistic changes to a document while largely retaining its original content. 
If an authorship representation avails of stylistic features, then we expect paraphrasing a document to {\em impair} its ability to match the document with other documents by the same author.

\subsection{Implementation details}\label{sec:paraphraseDetails}

To generate paraphrases, we use the STRAP paraphraser developed by \citet{Krishna2020}, which consists of a fine-tuned GPT-2 language model trained on \textsc{ParaNMT-50M}~\citep{wieting-gimpel-2018-paranmt}, a large dataset consisting of paraphrases of English sentences.

Because automatic paraphrasing models provide no guarantee that the proposed paraphrases of a document retain its meaning, we need to check this explicitly.
For this purpose we adopt the \textsc{BERTScore}~\citep{bertscore} as our primary similarity metric, rescaled to the unit interval. 
Unlike $n$-gram-matching metrics like \textsc{Bleu}, \textsc{BERTScore} leverages contextual embeddings to make predictions about semantic similarity. 

Because \textsc{STRAP} acts on {\em sentences} rather than {\em episodes}, we apply it independently to each sentence comprising an episode, with the following caveat.
Preliminary experiments revealed that the degree to which automatic paraphrasing retained meaning varied widely,
an issue that we mitigate as follows.
For each of the sixteen constituent documents $x$ of an episode, we paraphrase the sentences within $x$ to obtain $x'$ and calculate the mean \textsc{BERTScore} of each sentence of $x$ with its paraphrase in $x'$. We discard the eight $x'$ with lowest mean \textsc{BERTScore} to form the paraphrased episode, and also drop the eight corresponding $x$ from the original episode for comparability.

\subsection{Impact of paraphrasing on ranking}\label{sec:para}

For each domain, we train an authorship representation in the usual way,
perform the primary ranking experiment described in~\autoref{sec:setup}, and repeat the experiment after paraphrasing all the queries in the manner described in~\autoref{sec:paraphraseDetails}.
In~\autoref{table:uar-sbert} we report the MRR for the original experiment (Orig), the MRR for the paraphrased variation (Para), and the change in performance ($\Delta$) for each domain.
To serve as a baseline, we repeat the entire experiment with the SBERT model
in place of the trained authorship representation, which is denoted by UAR in~\autoref{table:uar-sbert}.

For each domain and each model, the MRR substantially decreased for the paraphrased queries relative to the original queries, which confirms that both models rely to some extent on author style. 
However, the performance degradation was much more pronounced for UAR than for SBERT,
suggesting that UAR captures style to a much greater extent than SBERT.

For each domain and each model, a paired $t$~test shows that
the decrease in MRR of the paraphrased queries relative to that of the original queries is significant with $p < 0.01$. 
Additionally, for each domain, a further $t$~test shows that the difference between the two models of the differences in MRR between the original and paraphrased queries is significant with $p<0.01$. 

\begin{table}[ht!]
\centering
\begin{tabular}{@{}clccc@{}}
\toprule
\bf Domain & \bf Model & \textbf{Orig} & \textbf{Para} & $\Delta$ \\ \midrule
\multirow{2}{*}{\bf fanfic} & \bf UAR & 0.325 & 0.139 & \textbf{0.186} \\
& \bf SBERT & 0.203 & 0.167 & 0.036 \\ \midrule
\multirow{2}{*}{\bf Reddit} & \bf UAR & 0.263 & 0.026 & \textbf{0.237} \\
& \bf SBERT & 0.043 & 0.026 & 0.017 \\ \midrule
\multirow{2}{*}{\bf Amazon} & \bf UAR & 0.266 & 0.025 & \textbf{0.241} \\
& \bf SBERT & 0.165 & 0.069 & 0.096 \\ \bottomrule
\end{tabular}
\caption{\small The impact on ranking performance of paraphrasing queries. Paraphrasing drastically impairs ranking performance of the UAR model relative to the baseline SBERT model, suggesting a reliance on stylistic features.}
\label{table:uar-sbert}
\vspace{-8pt}
\end{table}

\begin{figure*}[ht!]
    \centering
    \includegraphics[scale=.3]{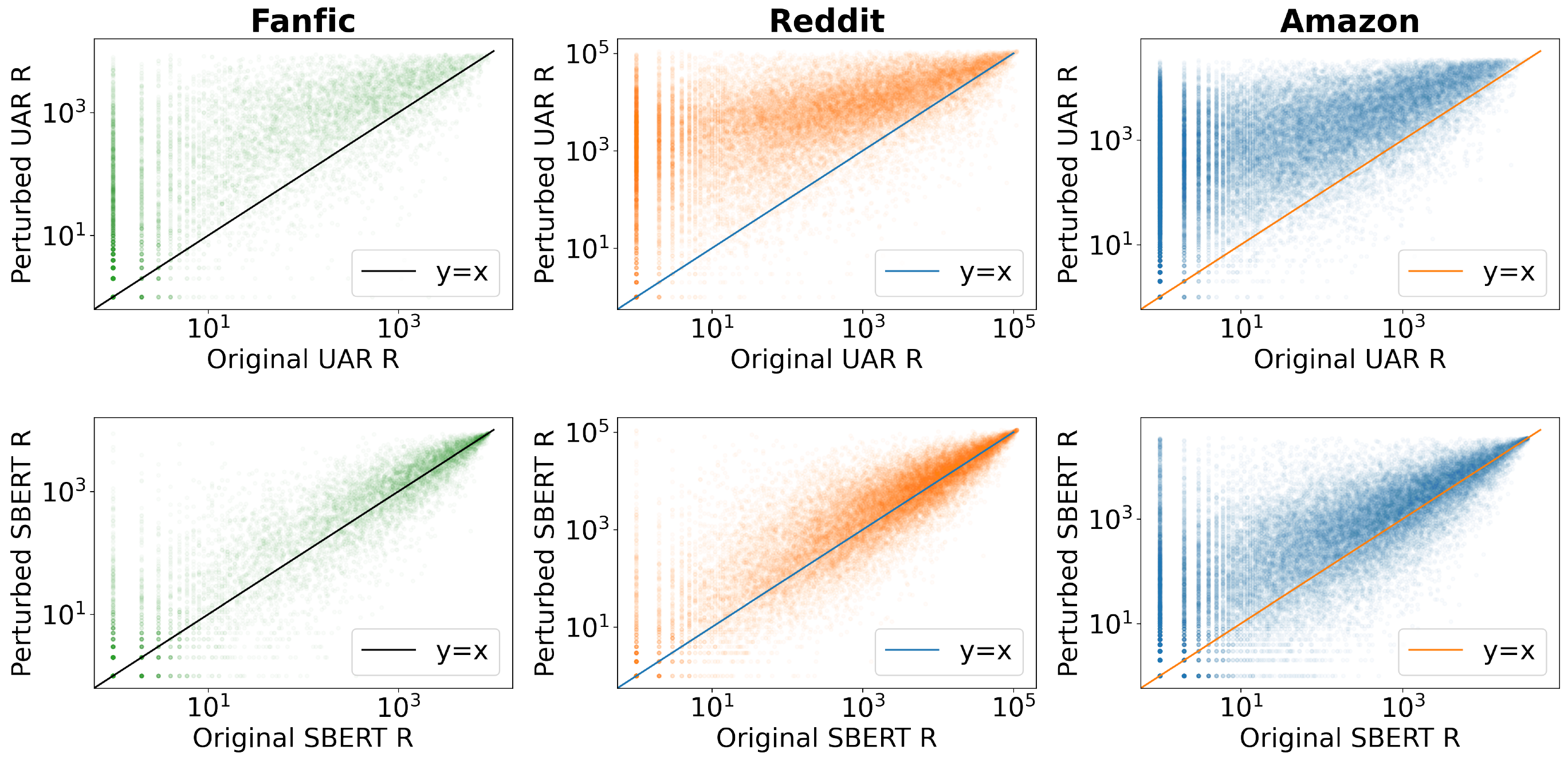}
    \caption{\small Rankings $r_i$ against $r_i'$. UAR has more points above the diagonal line $r=r'$ than SBERT, which correspond with queries for which paraphrasing hurts ranking performance.}
    \label{fig:rankvrank}
\vspace{-8pt}
\end{figure*}

We also present the results of this experiment in a more qualitative way in~\autoref{fig:rankvrank}.
Recall from~\autoref{sec:mrrAppendix} that for each query $q_i$ and its corresponding target $t_i$, our primary ranking experiment entails ranking all the targets $t_1,t_2,\ldots,t_N$ according to their similarity to $q_i$ and reporting the position $r_i$ of $t_i$ in the ranked list.
In \autoref{fig:rankvrank} we plot $r_i$ against $r_i'$ for each $1\le i\le M$, where $r'_i$ is the position of $t_i$ in the list ranked according to similarity to $q_i'$, the paraphrase of $q_i$.

Examples for which $r_i\approx r_i'$ correspond to points near the diagonal line shown, whereas examples for which $r_i>r_i'$ correspond to points above this line.
Note that for the UAR model, most points lie above the diagonal line, while for SBERT, the points are more evenly distributed across both sides of the line.
This again suggests that paraphrasing impairs the ranking performance of UAR much more dramatically than that of SBERT.

\subsection{Quality of paraphrases} \label{section:qual}
If the premise that our paraphrases retain meaning but alter style were not satisfied, then we would not be able to infer that paraphrasing the queries in~\autoref{sec:para} is responsible for the drop in ranking performance.
To confirm that the premise is largely satisfied, we propose the following metrics, both averaged over all the sentences comprising a query.
To assess the degree of content preservation between a query $q$ and its paraphrase $q'$, we calculate their \textsc{BERTScore}.
To confirm that $q'$ significantly modifies the style of $q$, rather than making only minor changes, we calculate their {\em normalized edit distance}.
While neither metric is perfect, together they provide a reasonable estimate of paraphrase quality.  

\begin{figure}[ht!]
    \centering
    \input{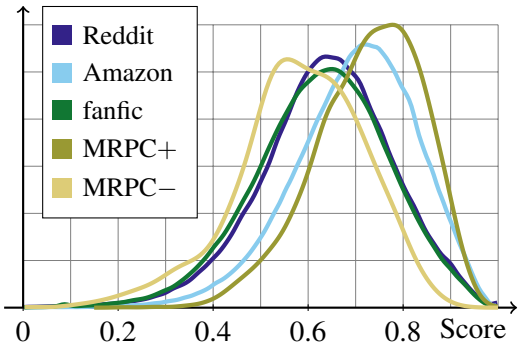}
    \caption{\small Distribution of \textsc{BERTScore}s comparing documents to their paraphrases.}
    \label{fig:sbert}
\end{figure}

\begin{figure}
    \centering
    \input{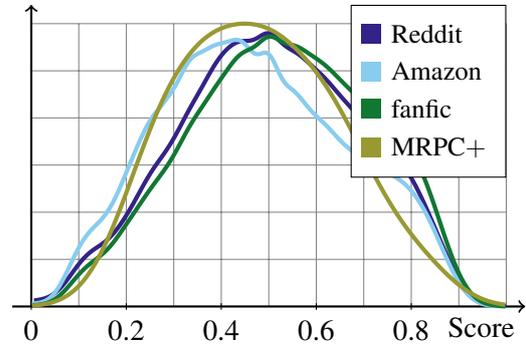}
    \caption{\small Distribution of edit distances between documents and their paraphrases.}
    \label{fig:edit}
\vspace{-8pt}
\end{figure}

As a baseline, we calculate the same metrics for the Microsoft Research Paraphrase Corpus (MRPC)~\citep{dolan2005automatically}.
\cref{fig:sbert,fig:edit} show that the distributions of both scores overlap substantially with those for the MRPC corpus restricted to sentence pairs deemed to constitute paraphrases by human annotators, which is labeled by MRPC$+$.
As a further baseline, \autoref{fig:sbert} also shows the distribution of MRPC scores restricted to pairs deemed {\em not} to constitute paraphrases, which is labeled by MRPC$-$.
Therefore we may rule out the possibility that the drop in ranking performance in~\autoref{sec:para} might be due to low-quality paraphrases.


\subsection{Impact of content overlap}
\begin{figure*}[ht!]
    \centering
    \includegraphics[scale=.3]{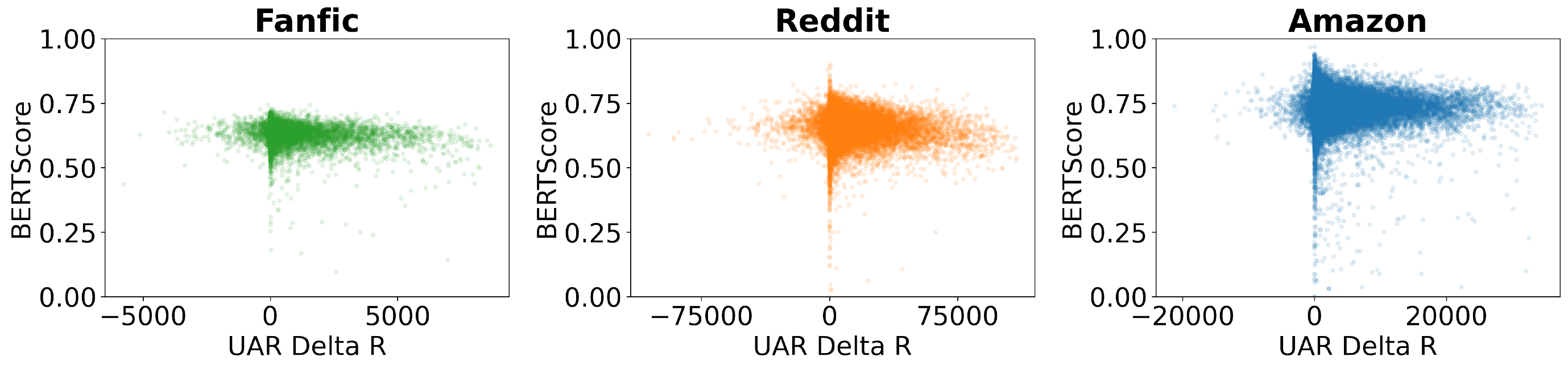}
    \caption{\small \textsc{BERTScore} against change in rank $\Delta R$. BERTScore is minimally correlated with $\Delta R$, suggesting that $\Delta R$ is not a function of content overlap.}
    \label{fig:correlations}
\vspace{-8pt}
\end{figure*}

As a final illustration, in~\autoref{fig:correlations} we plot the \textsc{BERTScore} $b_i$ of $q_i$ with $q_i'$ against the change in rank $\Delta r_i = r_i'-r_i$ for all $1\le i\le M$. If authorship representations were significantly influenced by content, then we might expect to see a strong negative relationship between $b_i$ and $\Delta r_i$. Instead, we observe little correlation, with Kendall's $\tau$ values of $-0.092$, $-0.019$, and $-0.015$ for the fanfic, Reddit, and Amazon domains respectively, suggesting that the ranking performance degradation in~\autoref{sec:para} cannot be well-explained by content overlap between $q_i$ and $q_i'$.

\subsection{A further application}\label{sec:further}
Although beyond the scope of this paper, we remark that a broader research problem is to determine whether the capacity of an authorship representation to encode style is {\em correlated} with its performance on the authorship attribution task. For example, if a new representation were introduced that performed better than UAR on attribution, would it necessarily encode style to a greater degree than UAR? Conversely, if a new approach were proposed to learn representations that encode style to a greater degree than UAR, would such representations perform better on attribution?

Addressing these questions will require assessing the degree to which a representation encodes style. We submit that the experiments presented in this paper are well-suited to making such assessments. As an illustration, we repeat the primary experiment described in~\autoref{sec:para} using two further instances of the UAR architecture introduced in~\autoref{sec:author_rep}, but trained on the Reddit histories of around 100K and 5M~authors respectively, in contrast with the version used throughout this paper, which was trained on the histories of around 1M~authors.\footnote{\label{foot:uar} We trained the smaller model with the dataset released by~\citet{andrews2019learning}. For the larger model we queried~\citet{baumgartner2020pushshift} for comments published between January 2015 and October 2019 by authors publishing at least 100 comments during that period. All three models were trained using the default hyperparameter settings of~\url{https://github.com/LLNL/LUAR}. The MRRs of UAR19, UAR, and UAR23 evaluated on a test set comprised of comments published future to those comprising the three training corpora are 0.482, 0.592, and 0.682 respectively.}

\begin{table}[ht!]
\centering
\begin{tabular}{@{}ccccc@{}}
\toprule
&\bf Training \\
\bf Model & \bf Examples & \textbf{Orig} & \textbf{Para} & $\Delta$ \\ \midrule
\bf UAR23 & 5M & 0.293 & 0.032 & 0.261\\
\bf UAR & 1M & 0.263 & 0.026 & 0.237 \\
\bf UAR19 & 100K & 0.188 & 0.019 & 0.169 \\\bottomrule
\end{tabular}
\caption{\small Impact of paraphrasing on attribution performance for authorship representations trained on varying numbers of Reddit users.}
\label{table:paraReddit}
\vspace{-8pt}
\end{table}
A paired $t$ test shows that the difference in rank induced by paraphrasing is significant with $p < 0.01$ for all three models. 
These differences are positively correlated with the MRR scores of the corresponding models, which are shown in~\autoref{foot:uar}, suggesting that improved attribution performance may be attributed at least in part to increased sensitivity to stylistic features.
\section{Generalization to Novel Tasks}\label{sec:gen}Our experiments have thus far focused on authorship prediction, a task which is presumably best addressed with a model capable of distinguishing writing styles.
We now use authorship representations to {\em directly} distinguish writing styles using a corpus of documents furnished with style annotations, namely the CDS dataset~\cite{Krishna2020}.
This consists of writings in disparate styles, including writings by two classical authors, namely Shakespeare and Joyce, historical American English from different eras, social media messages, lyrics, poetry, transcribed speech, and the Bible. With the notable exception of the two classical authors, most styles in CDS are not author-specific, but rather, represent broad stylistic categories. This means that identifying CDS styles is not the same problem as authorship prediction, an important observation we revisit below.

In addition, we repeat the experiment with a corpus furnished instead with {\em topic} annotations, namely the Reuters21578 dataset~\cite{lewis1997reuters}.
This is a popular benchmark in text classification consisting of financial news articles, each annotated by one or more topics. We note that certain topics may be associated with particular authors and editors, and therefore style could be a spurious correlate, although we nevertheless expect the authorship representation to perform worse \emph{relative} to the semantic baseline described below.

For each corpus, the experiment consists of simply applying an authorship representation trained on the Reddit dataset to two randomly chosen documents from the corpus. We used Reddit because it has been shown to yield representations that generalize well to other domains~\cite{Soto2021LearningUA}.
We record the dot product of the resulting vectors and pair this score with a binary indicator specifying whether the two documents carry the same labels.
Noting that predicting the binary indicator from the dot product is a highly imbalanced problem, with most document pairs bearing {\em non-matching} rather than {\em matching} labels, we simply construct the corresponding receiver operating characteristic (ROC) curve, an illustrative device intended to explore the tradeoffs in making that prediction by thesholding.
We report the equal error rate (EER), a simple summary statistic of the ROC curve. Smaller values of this metric are preferable. For good measure, we also report the area under that curve (AUC) in~\autoref{sec:aucAppendix}, another summary statistic of the ROC curve.

Finally, because writing style may be difficult to assess without sufficient text content, we also vary the amount of text contributing to the dot products mentioned above. Specifically, rather than predicting whether the label of a {\em single} document matches that of another on the basis of the dot product of their representations, we more generally predict whether a {\em group} of randomly chosen documents of the same label shares that label with another group of randomly chosen documents sharing another label on the basis of the dot product of the means over each group of the representations of their constituents.

As a baseline we repeat both experiments using the general purpose document embedding SBERT in place of the authorship representation. SBERT is commonly regarded as a semantic embedding, but is not typically used to discriminate writing styles without further training.

The rationale for the experiment is the following.
If the authorship representation primarily encodes stylistic features, then we would expect \emph{poor performance} relative to SBERT on the topic discrimination task since the task presumably does not involve stylistic distinction. However, we would expect {\em better performance} from the authorship representation than SBERT on the style discrimination task.

These expectations are borne out in the results reported in~\cref{fig:reuters,fig:style}, which show EER against the number of input documents for each task and each model.
The generalization performance of UAR on these novel tasks relative to SBERT is consistent with a representation that is sensitive to stylistic information. 
Namely, SBERT consistently outperforms UAR on topic classification, while UAR consistently outperforms SBERT on style classification.
We present 95\% confidence intervals for each curve as lighter regions of the same color surrounding the curve.
Although these were calculated using a bootstrap approach, the confidence intervals of the corresponding AUC results shown in~\cref{fig:reutersAUC,fig:styleAUC} of~\autoref{sec:aucAppendix} were calculated using a bootstrap-free calculation.

Also shown in~\crefrange{fig:reuters}{fig:styleAUC} are the results of the same experiments using the two variations of UAR introduced in~\autoref{sec:further}.
These additional models were included to support an auxiliary argument raised in~\autoref{sec:discussion}, but also afford an interesting but subtle insight about the current task.
Namely, although UAR19 performs {\em strictly worse} and UAR23 {\em strictly better} than UAR on the authorship attribution task, classifying style into broad categories is a different problem than authorship attribution, the latter dealing with fine-grained stylometric features. This accounts for the seemingly contradictory results in~\autoref{fig:style}, in which UAR19 performs {\em better} than UAR, which in turn performs better than UAR23. Indeed, training UAR on {\em more} authors produces representations that are more discriminative of individual authors, something which is at odds with identifying broad stylistic categories for the simple reason that being exposed to more authors affords more opportunities for UAR to discover stylistic features that distinguish authors.

Notwithstanding these observations, we remark that within the CDS dataset, certain styles are likely to be correlated with particular topics, while in the Reuters dataset, certain authors are likely to often write about particular topics. This would suggest that SBERT might perform better on CDS and UAR better on Reuters than one might expect, so the absolute performance on both tasks is not particularly informative.

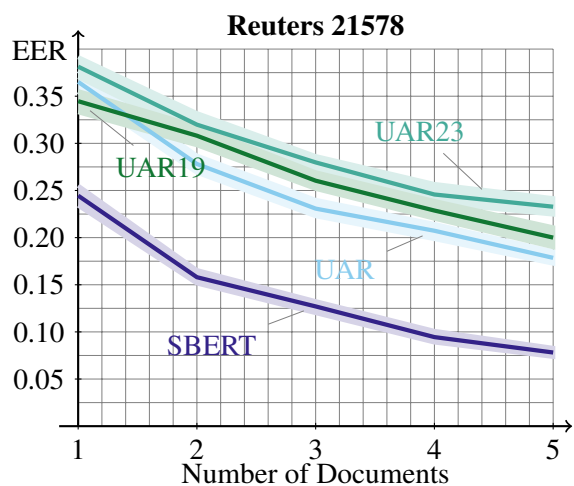
\begin{figure}[ht!]
    \centering
    \begin{tikzpicture}[scale=1.25,pin distance=0.5cm]
\draw[xstep=0.25,ystep=0.25,gray,very thin] (0,0) grid (5,4);
\draw[thick,->] (-0.2,0) -- (5.2,0);
\draw (2.5,4.25) node {\bf Reuters 21578};
\draw (2.5,-.5) node {Number of Documents};
\draw[thick,->] (0,0) -- (0,4.2) node[below left] {EER};
\foreach \x in {1,2,3,4,5}{
  \def\X{5*\x/4-5/4}
  \draw (\X,1pt) -- (\X,-1pt) node[below] {\x};
}
\foreach \y/\l in {0.05,0.10,0.15,0.20,0.25,0.30,0.35}
  \draw (1pt,10*\y) -- (-1pt,10*\y) node[left] {\y};
\definecolor{c0}{RGB}{51,34,136}
\definecolor{c1}{RGB}{136,204,238}
\definecolor{c2}{RGB}{68,170,153}
\definecolor{c3}{RGB}{17,119,51}
\definecolor{c4}{RGB}{153,153,51}

\draw[c0!20,fill=c0!20,ultra thick]
(0.000,2.5507) -- (1.250,1.6551) -- (2.500,1.3256) -- (3.750,1.0101) -- (5.000,0.8300) -- (5.000,0.7359) -- (3.750,0.8883) -- (2.500,1.2002) -- (1.250,1.5142) -- (0.000,2.3401);

\draw[c1!20,fill=c1!20,ultra thick]
(0.000,3.7679) -- (1.250,2.8695) -- (2.500,2.3994) -- (3.750,2.1443) -- (5.000,1.8535) -- (5.000,1.7246) -- (3.750,1.9855) -- (2.500,2.2280) -- (1.250,2.6893) -- (0.000,3.5448);

\draw[c2!20,fill=c2!20,ultra thick]
(0.000,3.9145) -- (1.250,3.3200) -- (2.500,2.8645) -- (3.750,2.5683) -- (5.000,2.4178) -- (5.000,2.2452) -- (3.750,2.3678) -- (2.500,2.7051) -- (1.250,3.0830) -- (0.000,3.7031);

\draw[c3!20,fill=c3!20,ultra thick]
(0.000,3.5531) -- (1.250,3.1821) -- (2.500,2.6921) -- (3.750,2.3879) -- (5.000,2.1087) -- (5.000,1.8994) -- (3.750,2.2019) -- (2.500,2.5178) -- (1.250,2.9804) -- (0.000,3.3306);

\draw[ultra thick,c0]
(0.000,2.4453) -- (1.250,1.5801) -- (2.500,1.2702) node[pin=200:SBERT]{} -- (3.750,0.9449) -- (5.000,0.7814);
\draw[ultra thick,c1]
(0.000,3.6518) -- (1.250,2.7811) -- (2.500,2.3077) -- (3.750,2.0729) node[pin=200:UAR]{} -- (5.000,1.7854);
\draw[ultra thick,c2]
(0.000,3.8138) -- (1.250,3.1984) -- (2.500,2.7978) -- (3.750,2.4575) -- node[pin=110:UAR23]{} (5.000,2.3279);
\draw[ultra thick,c3]
(0.000,3.4471) node[pin=300:UAR19]{} -- (1.250,3.0810) -- (2.500,2.6038) -- (3.750,2.2874) -- (5.000,2.0000);

\end{tikzpicture}
    \caption{\small Equal error rate (EER) for UAR and SBERT on topic distinction as the size of the writing sample is varied. Smaller values of EER correspond with better performance.}
    \label{fig:reuters}
\end{figure}
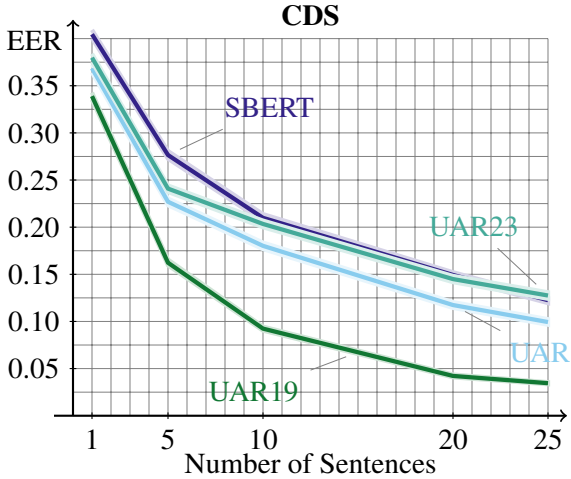
\begin{figure}
    \centering
    \begin{tikzpicture}[scale=1.25,pin distance=0.5cm]
\draw[xstep=0.2,ystep=0.25,gray,very thin] (0,0) grid (5,4);
\draw[thick,->] (-0.2,0) -- (5.2,0); 
\draw (2.5,4.25) node {\bf CDS};
\draw (2.5,-.5) node {Number of Sentences};
\draw[thick,->] (0,0) -- (0,4.2) node[below left] {EER};
\foreach \x in {1, 5, 10, 20, 25}
  \draw (\x/5,1pt) -- (\x/5,-1pt) node[below] {\x};
\foreach \y/\l in {0.05,0.10,0.15,0.20,0.25,0.30,0.35}
  \draw (1pt,10*\y) -- (-1pt,10*\y) node[left] {\y};

\definecolor{c0}{RGB}{51,34,136}
\definecolor{c1}{RGB}{136,204,238}
\definecolor{c2}{RGB}{68,170,153}
\definecolor{c3}{RGB}{17,119,51}
\definecolor{c4}{RGB}{153,153,51}
\draw[c0!20,fill=c0!20,ultra thick]
(0.200,4.0924) -- (1.000,2.8072) -- (2.000,2.1383) -- (4.000,1.5218) -- (5.000,1.2498) -- (5.000,1.1949) -- (4.000,1.4650) -- (2.000,2.0636) -- (1.000,2.7260) -- (0.200,4.0040);
\draw[ultra thick,c0]
(0.200,4.0463) -- (1.000,2.7651) node[pin=30:SBERT]{} -- (2.000,2.0987) -- (4.000,1.4931) -- (5.000,1.2251);

\draw[c1!20,fill=c1!20,ultra thick]
(0.200,3.7255) -- (1.000,2.3029) -- (2.000,1.8364) -- (4.000,1.2048) -- (5.000,1.0239) -- (5.000,0.9654) -- (4.000,1.1437) -- (2.000,1.7676) -- (1.000,2.2328) -- (0.200,3.6416);
\draw[ultra thick,c1]
(0.200,3.6853) -- (1.000,2.2694) -- (2.000,1.8019) -- (4.000,1.1740) node[pin=330:UAR]{} -- (5.000,0.9937);

\draw[c2!20,fill=c2!20,ultra thick]
(0.200,3.8398) -- (1.000,2.4418) -- (2.000,2.0696) -- (4.000,1.4829) -- (5.000,1.3093) -- (5.000,1.2404) -- (4.000,1.4170) -- (2.000,1.9991) -- (1.000,2.3738) -- (0.200,3.7543);
\draw[ultra thick,c2]
(0.200,3.7981) -- (1.000,2.4090) -- (2.000,2.0341) -- (4.000,1.4494) -- (5.000,1.2751) node[pin=110:UAR23]{} ;

\draw[c3!20,fill=c3!20,ultra thick]
(0.200,3.4276) -- (1.000,1.6559) -- (2.000,0.9447) -- (4.000,0.4371) -- (5.000,0.3555) -- (5.000,0.3343) -- (4.000,0.4072) -- (2.000,0.9029) -- (1.000,1.5948) -- (0.200,3.3517);
\draw[ultra thick,c3]
(0.200,3.3897) -- (1.000,1.6230) -- (2.000,0.9239) -- node[pin=200:UAR19]{} (4.000,0.4221) -- (5.000,0.3451);

\end{tikzpicture}
    \caption{\small Equal error rate (EER) for UAR and SBERT on style distinction as the size of the writing sample is varied. Smaller values of EER correspond with better performance.}
    \label{fig:style}
\vspace{-8pt}
\end{figure}

\section{Discussion}\label{sec:discussion}\paragraph{Findings}
We have examined properties of an exemplary authorship representation construction,
finding consistent evidence that the success of the representations it engenders at distinguishing authors may be attributed in large part to their sensitivity to style.
First, the masking experiments of~\autoref{sec:mask} show that for sufficiently large training corpora, masking a large fraction of content words at training time does not significantly affect ranking performance on held-out data, suggesting that these representations are largely invariant to the presence of content words.
On the other hand, the paraphrasing experiments of~\autoref{sec:andrew}, which seek to alter writing style while preserving content, confirm that paraphrasing drastically impairs ranking performance. Taken together, these experiments suggest that the authorship representations considered are  indeed sensitive to stylistic features.
This conclusion is corroborated in~\autoref{sec:gen} where we see poor generalization of one of these representations to a topic discrimination task, but better generalization to a style discrimination task, both assessments relative to a semantic baseline.

\paragraph{Limitations}
All of the experiments in this paper involve instances of the UAR construction.
Since our primary research question involves testing the capacity of representations trained for the authorship prediction task to capture stylistic features, we select this construction because there is prior evidence that the representations it engenders perform well at zero-shot cross-domain transfer, for certain combinations of source and target domains, which likely requires some degree of stylistic sensitivity~\cite{Soto2021LearningUA}.

By design, our analysis is focused on aggregate model behavior. While this addresses the high-level research questions we pose in the introduction, such \emph{global} analysis does not enable predictions about which specific \emph{local} features are involved in model predictions. As such, an important avenue for future work lies in developing methods that can faithfully identify local authorship features. To this end, frameworks for evaluating the quality of explanations, such as~\citet{pruthi2022evaluating}, are essential. Beyond the usual challenges of explaining the decisions of deep neural networks, explaining author style may pose further challenges, such as the need to identify groups of features that in combination predict authorship.

We emphasize that completely disentangling style from content may not be attainable, since certain aspects of writing likely blur the line between style and content \citep{jafaritazehjani2020}.
For example, we notice degradation in ranking performance of the SBERT model in \autoref{table:uar-sbert},
suggesting that to some extent, SBERT features are also stylistic.
Nonetheless, UAR exhibits a markedly larger degradation in performance, suggesting a greater degree of sensitivity to writing style.

\paragraph{Broader Impact}
This work contributes to the broader goal of formulating notions of {\em content} and {\em style} that constitute mutually exclusive and collectively exhaustive aspects of writing that may be embedded in orthogonal subspaces of a Euclidean space. Not knowing whether this ambition is fully realizable, but hopeful others will explore the question in future research, we resign ourselves in this paper to exhibiting two embeddings that accomplish the objective to a limited extent. 
Specifically, we focus on UAR, which we show to mostly encode style, and SBERT, which is widely assumed to encode content.
This being an imperfect decomposition, the primary goal of this paper is to qualitatively assess the {\em degree} to which UAR encodes style rather than content.

Authorship attribution is likely to be a task that benefits from a representation that is relatively stable over time, specifically an encoding capturing primarily writing style.
To this end, another open question is whether a representation may be constructed that encodes style to a {\em greater} degree than UAR, and if so, whether the representation improves performance on the attribution task.
If such a representation were proposed in the future, the experiments we propose in this paper could be used to validate the assertion that it encodes style to a greater degree than UAR.

On the other hand, content features constitute perfectly legitimate discriminators of authorship in some cases. For example, an author who discusses only a narrow range of topics on a particular forum may easily be distinguished from other authors on the basis of topic features.
Not knowing under which circumstances and to what degree content plays a role in authorship attribution, we maintain that the relationship between the performance of a representation on the attribution task and the degree to which the representation encodes content should be explored fully, something that will again require an estimate of the degree to which a representation encodes style.

Another promising application of authorship representations is {\em style transfer}, where one hopes to rephrase a given statement in the style of a given author. This has been analogously accomplished in the domain of speech, resulting in the ability to have a given statement recited in the voice of a given speaker~(see e.g.\ \citet{kim2021conditional}). The primary ingredient in this task is a speaker embedding, which is analogous to an authorship representation. However, by construction, a speaker embedding encodes almost exclusively {\em acoustic} features, but encodes content features, namely the specific words spoken, to a very limited degree. The fact that this observation might be the primary reason for the success of speech transfer portends possible difficulties for the style transfer task. However, as with authorship attribution, the relationship between the success of using a representation for style transfer and the degree to which the representation encodes style should also be fully explored, and again, the experiments proposed in this paper would constitute a natural assessment of that degree.

\section*{Acknowledgments}We thank the TACL reviewers and action editors for their insightful comments. We also thank Carina Kauf for the initial masking idea and Hope McGovern for early discussions on PertLE. Part of this work was performed under the auspices of the US Department of Energy by Lawrence Livermore National Laboratory under Contract DEAC52-07NA27344.
\bibliography{main}
\bibliographystyle{acl_natbib}
\appendix
\section{Appendix}
\subsection{Further dataset details}\label{sec:dataAppendix}
The experiments in this paper involve the same datasets used by~\citet{Soto2021LearningUA}. 
These datasets consist of Reddit comments~\citep{andrews2019learning,baumgartner2020pushshift}, Amazon product reviews~\citep{ni2019justifying}, and fanfiction short stories~\citep{PAN2020}, all organized according to author and sorted by publication time.
\autoref{tab:dataset} presents some statistics of each dataset.
Because these are all {\em anonymous} domains, we use account names as a stand-in for author labels, as proposed in the papers cited above. We recognize that this recourse may introduce a small amount of label noise, since an author may operate multiple accounts, or an account may contain contributions of multiple authors, both of which we assume to be relatively rare.

Each domain is composed of two disjoint splits used independently to train and evaluate models.
In the case of Amazon and Reddit, the documents comprising the evaluation split
were published in the {\em future}
relative to those comprising the training split.
In addition, in the fanfic domain, the {\em authors} contributing to the evaluation split 
are disjoint from those contributing to the training split.

We use a dataset derived from the Weibo social network in~\autoref{sec:tertle}.
This dataset contains posts published in the year 2012, primarily in Chinese.
We use the first 26 weeks for training, the next 13 weeks for validation, and the final 13 weeks for evaluation.
Restricting to authors who have posted a minimum of 50 times and a maximum of 1,000 times results in 94,292 authors in the training split and 90,489 in the evaluation split.

\begin{table}
\begin{center}
\begin{tabular}{crrc}
& \bf Train & \bf Test & \bf Documents \\
\bf Domain & \bf Authors & \bf Authors & \bf per Author \\\toprule
Amazon & 100K & 35K & $\ge 100$ \\
fanfic & 41K & 16K & $\ge 2$ \\
Reddit & 120K & 121K & $\ge 100$ \\
Weibo & 94K & 90K & $\ge 50$
\\\bottomrule
\end{tabular}
\end{center}
\caption{\small Dataset statistics.}
\label{tab:dataset}
\end{table}
\subsection{More on evaluation}\label{sec:mrrAppendix}
For each training domain we train an authorship representation $f_\theta$,
which maps episodes to the unit sphere in $\mathbb{R}^{512}$.
In fact, we independently train {\em three} such representations for each training domain
in an effort to reduce variance, a detail we revisit below after discussing
the calculation of {\em mean reciprocal rank} (MRR) for a single representation $f_\theta$.

We compare the authorship of two episodes through the dot product
of their images under $f_\theta$, which range from $+1$ to $-1$,
with $+1$ (respectively $-1$) corresponding to the strongest
prediction that the two input episodes were (respectively, were {\em not})
composed by the same author.

To calculate the MRR, we evaluate $f_\theta$ on all the episodes
of each evaluation corpus.
Each evaluation corpus consists of episodes
$q_1,q_2,\ldots,q_M,t_1,t_2,\ldots,t_N$ for some $M\le N$,
where $t_1,t_2,\ldots,t_N$ were each composed by a distinct author 
and where $q_i$ and $t_i$ have the same author for all $1\le i\le M$.

For each $1\le i\le M$ we sort the vectors
$f_\theta\left(t_1\right), f_\theta\left(t_2\right),\ldots, f_\theta\left(t_N\right)$
according to their dot products with $f_\theta\left(q_i\right)$,
with those $f_\theta\left(t_j\right)$ with the {\em greatest} dot products
having the {\em lowest} numbered positions in the ranked list.
We denote the position of $f_\theta\left(t_i\right)$ in this list by $r_i\left(\theta\right)$
and define
$\mathop{\mathrm{MRR}}\left(\theta\right)=\frac{1}{M}\sum_{i=1}^M\frac{1}{r_i\left(\theta\right)}$.

Finally, as mentioned above, to reduce variance, we independently train {\em three} representations
$f_{\theta_1}, f_{\theta_2}, f_{\theta_3}$ for each domain
and report the mean $\frac{1}{3}\sum_{k=1}^3\mathop{\mathrm{MRR}}\left(\theta_k\right)$
for each evaluation domain.
\subsection{The TertLE schema} \label{sec:tertle} 

In addition to being expensive to compute, POS tags are unavailable in many low-resource languages.
A more quantitative observation about the distinction between content and function words is that function words, such as~{\em the}, tend to be very frequent in a given language, while content words, such as~{\em wallpaper}, tend to be infrequent overall, but may be relatively frequent in documents dealing with those topics. In other words, content words may have higher Term Frequency-Inverse Document Frequency (TF-IDF) scores than function words.
One may interpret the highest-scoring words in a document as the most unique or relevant to that document and thus the most likely to be content words.
Based on this observation, we explore the possibility of masking words according to their TF-IDF scores rather than their POS tags, an approach we call {\em TertLE}.

For each domain we fit a TF-IDF model to the training split and use it to index all the documents in the corpus.
We introduce the {\em TertLE Grande} and {\em TertLE Lite} levels, in which we mask the top-scoring proportion~$p$ of words in each document, where $p$ is the proportion of words masked in the same domain by the corresponding PertLE Grande or PertLE Lite schema respectively. These values of $p$ are shown in~\autoref{table:percentages}. 

The experiment proceeds exactly as in~\autoref{sec:pertle} with results shown in~\autoref{table:tertle}.
Each number reported is the mean MRR computed by three independently-trained models, where 0.011 is the maximum sample standard deviation over all experiments reported in the table.

We observe that the Lite model outperforms the unmasked model in most cases and is also generally better than the Grande model, especially in the fanfic domain. Once again, the Grande model is generally worse than the unmasked model, but only slightly, and even improves on the unmasked model in some settings. If words with high TF-IDF scores are indeed primarily content words, then this experiment again suggests that authorship representations rely little on content and more heavily on writing style.

\begin{figure}[th!]
\begin{center}
\definecolor{c0}{RGB}{51,34,136}
\definecolor{c1}{RGB}{136,204,238}
\definecolor{c2}{RGB}{68,170,153}
\definecolor{c3}{RGB}{17,119,51}
\definecolor{c4}{RGB}{153,153,51}
\begin{tikzpicture}[scale=1.5]

\draw[xstep=0.3,ystep=0.25,gray,very thin] (0,-1) grid (2.1,8.75);
\draw (-1.95,-0.125) -- (2.1,-0.125);
\draw (-1.95,2.875) -- (2.1,2.875);
\draw (-1.95,5.875) -- (2.1,5.875);

\foreach \l/\y/\z in { Weibo/-1/-0.25, fanfic/0.00/2.75, Amazon/3/5.75, Reddit/6/8.75 }
  \draw[white] (-1.95,\y) rectangle (-1.3,\z) node[pos=0.5,black] {\begin{sideways}\l\end{sideways}};

\foreach \l/\y/\z in {
Weibo/-1/-0.25,
fanfic/0/0.75, Amazon/1/1.75, Reddit/2/2.75,
fanfic/3/3.75, Amazon/4/4.75, Reddit/5/5.75,
fanfic/6/6.76, Amazon/7/7.75, Reddit/8/8.75 }
  \draw[white] (-1.3,\y) rectangle (-0.65,\z) node[pos=0.5,black] {\begin{sideways}\l\end{sideways}};

\foreach \x/\c/\y/\l in {
0.416/c1/00.00/L, 0.306/c2/00.25/G, 0.434/c3/00.50/U,
0.277/c1/01.00/L, 0.104/c2/01.25/G, 0.205/c3/01.50/U,
0.085/c1/02.00/L, 0.040/c2/02.25/G, 0.069/c3/02.50/U,
0.257/c1/03.00/L, 0.252/c2/03.25/G, 0.219/c3/03.50/U,
0.142/c1/05.00/L, 0.144/c2/05.25/G, 0.123/c3/05.50/U,
0.267/c1/06.00/L, 0.253/c2/06.25/G, 0.258/c3/06.50/U,
0.459/c1/07.00/L, 0.460/c2/07.25/G, 0.452/c3/07.50/U,
0.342/c1/08.00/L, 0.345/c2/08.25/G, 0.360/c3/08.50/U
}{
  \draw[white] (-0.65,\y) rectangle (0,\y+0.25) node[pos=0.5,black]{\l};
  \draw[fill=\c] (0,\y) rectangle (\x*3,\y+0.25);
  \draw[white] (\x*3+.1,\y) rectangle (\x*3+0.5,\y+0.25) node[pos=0.5,black]{\x};
}

\foreach \x/\c/\y/\l in {
0.637/c1/04.00/L, 0.615/c2/04.25/G, 0.636/c3/04.50/U,
0.537/c1/-1.00/L, 0.434/c2/-0.75/G, 0.559/c3/-0.50/U
}{
  \draw[white] (-0.65,\y) rectangle (0,\y+0.25) node[pos=0.5,black]{\l};
  \draw[fill=\c] (0,\y) rectangle (\x*3,\y+0.25);
  \draw[\c] (\x*3-0.6,\y) rectangle (\x*3,\y+0.25) node[pos=0.5,white]{\x};
}

\draw[thick] (-1.95,8.875) -- (2.1,8.875);
\draw[white] (-1.95,9) rectangle (-1.3,9.25) node[pos=0.5,black] {\bf Train};
\draw[white] (-1.3,9) rectangle (-0.65,9.25) node[pos=0.5,black] {\bf Test};
\draw[white] (-.65,9) rectangle (0,9.25) node[pos=0.5,black] {\bf Level};
\draw[white] (0,9) rectangle (2.1,9.25) node[pos=0.5,black] {\bf MRR};

\end{tikzpicture}
\caption{\small TertLE MRR results for models trained on either unmasked data (U) or data masked according to the TertLE Grande (G) or the TertLE Lite (L) schema.}
\label{table:tertle}
\end{center}
\end{figure}

As mentioned above, one advantage of the TertLE schema is that it obviates POS tagging.
To illustrate this potential, and to determine whether similar patterns hold in another language, we repeat our TertLE experiment with a Chinese dataset scraped from the Weibo social network. See~\autoref{sec:dataAppendix} for more details on this dataset.
This requires replacing the SBERT component of the UAR architecture with a Chinese BERT pre-trained using whole word masking~\citep{chinese-wwm}.  
The results of the experiment with the Weibo dataset, displayed in~\autoref{table:tertle}, show an overall pattern similar to that of the English-focused experiments.

\begin{table}
\begin{center}
\begin{tabular}{cccc}
&&\multicolumn{2}{c}{\bf Level}\\\cmidrule{3-4}
&&  G & L \\\toprule
\multirow{4}{*}{\begin{sideways}\bf Domain \end{sideways}} 
& \bf Reddit &  0.438 & 0.177 \\
& \bf Amazon &  0.445 & 0.186 \\
& \bf Fanfic &  0.425 & 0.147 \\
& \bf Weibo &  0.436 & 0.170 \\
\bottomrule
\end{tabular}
\caption{\small Proportions of tokens masked by the Grande (G) or Lite (L) levels of both PertLE and TertLE.
For Weibo each proportion is the mean of the proportions at the same level in the other three domains.
}
\label{table:percentages}
\end{center}
\end{table}

\subsection{Further AUC Results}\label{sec:aucAppendix}
\begin{figure}
    \centering
    \begin{tikzpicture}[scale=1.25,pin distance=0.5cm]
\draw[xstep=0.25,ystep=0.25,gray,very thin] (0,0) grid (5,4);
\draw[thick,->] (-0.2,0) -- (5.2,0);
\draw (2.5,4.25) node {\bf Reuters 21578};
\draw (2.5,-.5) node {Number of Documents};
\draw[thick,->] (0,0) -- (0,4.2) node[below left] {AUC};
\foreach \x in {1,2,3,4,5}{
  \def\X{5*\x/4-5/4}
  \draw (\X,1pt) -- (\X,-1pt) node[below] {\x};
}
\foreach \y/\l in {0.65,0.70,0.75,0.80,0.85,0.90,0.95}
  \draw (1pt,10*\y-6) -- (-1pt,10*\y-6) node[left] {\y};
\definecolor{c0}{RGB}{51,34,136}
\definecolor{c1}{RGB}{136,204,238}
\definecolor{c2}{RGB}{68,170,153}
\definecolor{c3}{RGB}{17,119,51}
\definecolor{c4}{RGB}{153,153,51}

\draw[c0!20,fill=c0!20,ultra thick]
(0.000,2.1472) -- (1.250,2.9923) -- (2.500,3.4219) -- (3.750,3.6257) -- (5.000,3.7069) -- (5.000,3.6333) -- (3.750,3.5482) -- (2.500,3.3194) -- (1.250,2.8314) -- (0.000,1.9196);

\draw[c1!20,fill=c1!20,ultra thick]
(0.000,0.8859) -- (1.250,1.9538) -- (2.500,2.5359) -- (3.750,2.7575) -- (5.000,3.0322) -- (5.000,2.9061) -- (3.750,2.6130) -- (2.500,2.3795) -- (1.250,1.7520) -- (0.000,0.6377);

\draw[c2!20,fill=c2!20,ultra thick]
(0.000,0.6999) -- (1.250,1.5178) -- (2.500,2.0577) -- (3.750,2.3358) -- (5.000,2.5696) -- (5.000,2.4122) -- (3.750,2.1626) -- (2.500,1.8745) -- (1.250,1.2988) -- (0.000,0.4497);

\draw[c3!20,fill=c3!20,ultra thick]
(0.000,1.0534) -- (1.250,1.7402) -- (2.500,2.2975) -- (3.750,2.5162) -- (5.000,2.7941) -- (5.000,2.6255) -- (3.750,2.3355) -- (2.500,2.1082) -- (1.250,1.5135) -- (0.000,0.7978);

\draw[ultra thick,c0]
(0.000,2.0334) -- (1.250,2.9118) node[pin=100:SBERT]{} -- (2.500,3.3706) -- (3.750,3.5869) -- (5.000,3.6701);
\draw[ultra thick,c1]
(0.000,0.7618) -- (1.250,1.8529) -- (2.500,2.4577) -- (3.750,2.6852)  node[pin=150:UAR]{} -- (5.000,2.9691);
\draw[ultra thick,c2]
(0.000,0.5748) -- (1.250,1.4083) -- (2.500,1.9661)  node[pin=300:UAR23]{} -- (3.750,2.2492) -- (5.000,2.4909);
\draw[ultra thick,c3]
(0.000,0.9256) node[pin=85:UAR19]{}  -- (1.250,1.6268) -- (2.500,2.2029) -- (3.750,2.4258) -- (5.000,2.7098);

\end{tikzpicture}
    \caption{\small Area under the ROC curve (AUC) for UAR and SBERT on topic distinction as the size of the writing sample is varied. Larger values of AUC correspond with better performance.}
    \label{fig:reutersAUC}
\end{figure}
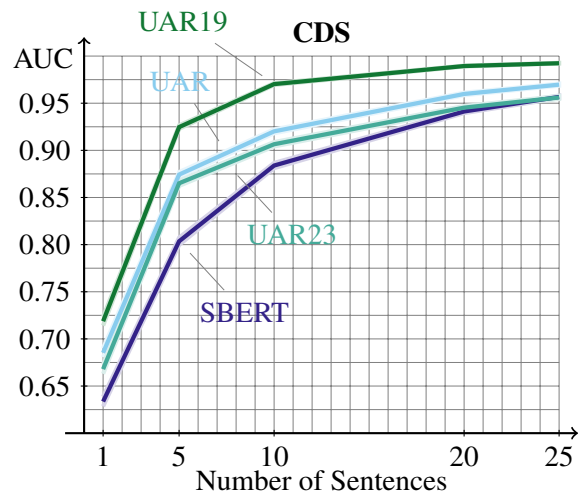
\begin{figure}
    \centering
    \begin{tikzpicture}[scale=1.25,pin distance=0.5cm]
\draw[xstep=0.2,ystep=0.25,gray,very thin] (0,0) grid (5,4);
\draw[thick,->] (-0.2,0) -- (5.2,0); 
\draw (2.5,4.25) node {\bf CDS};
\draw (2.5,-.5) node {Number of Sentences};
\draw[thick,->] (0,0) -- (0,4.2) node[below left] {AUC};
\foreach \x in {1, 5, 10, 20, 25}
  \draw (\x/5,1pt) -- (\x/5,-1pt) node[below] {\x};
\foreach \y/\l in {0.65,0.70,0.75,0.80,0.85,0.90,0.95}
  \draw (1pt,10*\y-6) -- (-1pt,10*\y-6) node[left] {\y};

\definecolor{c0}{RGB}{51,34,136}
\definecolor{c1}{RGB}{136,204,238}
\definecolor{c2}{RGB}{68,170,153}
\definecolor{c3}{RGB}{17,119,51}
\definecolor{c4}{RGB}{153,153,51}

\draw[c0!20,fill=c0!20,ultra thick]
(0.200,0.3859) -- (1.000,2.0733) -- (2.000,2.8663) -- (4.000,3.4295) -- (5.000,3.5804) -- (5.000,3.5529) -- (4.000,3.3952) -- (2.000,2.8114) -- (1.000,1.9972) -- (0.200,0.2850);

\draw[c1!20,fill=c1!20,ultra thick]
(0.200,0.9003) -- (1.000,2.7693) -- (2.000,3.2217) -- (4.000,3.6112) -- (5.000,3.7075) -- (5.000,3.6861) -- (4.000,3.5848) -- (2.000,3.1813) -- (1.000,2.7163) -- (0.200,0.8052);

\draw[c2!20,fill=c2!20,ultra thick]
(0.200,0.7284) -- (1.000,2.6786) -- (2.000,3.0876) -- (4.000,3.4699) -- (5.000,3.5754) -- (5.000,3.5461) -- (4.000,3.4358) -- (2.000,3.0419) -- (1.000,2.6232) -- (0.200,0.6298);

\draw[c3!20,fill=c3!20,ultra thick]
(0.200,1.2337) -- (1.000,3.2651) -- (2.000,3.7121) -- (4.000,3.8984) -- (5.000,3.9274) -- (5.000,3.9209) -- (4.000,3.8900) -- (2.000,3.6933) -- (1.000,3.2282) -- (0.200,1.1413);

\draw[ultra thick,c0]
(0.200,0.3354) -- (1.000,2.0352)  node[pin=280:SBERT]{} -- (2.000,2.8389) -- (4.000,3.4124) -- (5.000,3.5667);
\draw[ultra thick,c1]
(0.200,0.8527) -- (1.000,2.7428) --  node[pin=95:UAR]{} (2.000,3.2015) -- (4.000,3.5980) -- (5.000,3.6968);
\draw[ultra thick,c2]
(0.200,0.6791) -- (1.000,2.6509) --  node[pin=280:UAR23]{} (2.000,3.0648)  -- (4.000,3.4529) -- (5.000,3.5608) ;
\draw[ultra thick,c3]
(0.200,1.1875) -- (1.000,3.2466) -- (2.000,3.7027)  node[pin=130:UAR19]{} -- (4.000,3.8942) -- (5.000,3.9241);
\end{tikzpicture}
    \caption{\small Area under the ROC curve (AUC) for UAR and SBERT on style distinction as the size of the writing sample is varied. Larger values of AUC correspond with better performance.}
    \label{fig:styleAUC}
\end{figure}
\cref{fig:reutersAUC,fig:styleAUC} report
the area under the receiver operating characteristic curves (AUC) of the experiments in~\autoref{sec:gen}. AUC is a further summary statistic of the ROC that in contrast to EER, admits a bootstrap-free confidence estimation.

\end{document}